\documentclass[a4paper,twoside]{article}
\usepackage{booktabs}           
\usepackage{multirow}           
\usepackage{amsmath}
\usepackage{amsfonts}           
\usepackage{amssymb}
\usepackage{caption}
\usepackage{subcaption}
\usepackage{hyperref}
\usepackage{graphicx}
\usepackage[T1]{fontenc}
\usepackage{authblk}
\usepackage[margin=3cm]{geometry}
\usepackage{fancyhdr}
\begin{document}
\title{Evaluating Link Prediction Explanations for Graph Neural Networks}
%
%
\author[1]{Claudio Borile}
\author[1]{Alan Perotti}
\author[1]{André Panisson}
\affil[1]{CENTAI Institute, Turin, Italy}

%
%
%
\maketitle              
\begin{abstract}
Graph Machine Learning (GML) has numerous applications, such as node/graph classification and link prediction, in real-world domains.
Providing human-understandable explanations for GML models is a challenging yet fundamental task to foster their adoption, 
but validating explanations for link prediction models has received little attention.

In this paper, we provide quantitative metrics to assess the quality of link prediction explanations, with or without ground-truth. State-of-the-art explainability methods for Graph Neural Networks are evaluated using these metrics.
We discuss how underlying assumptions and technical details specific to the link prediction task, such as the choice of distance between node embeddings, can influence the quality of the explanations.

\end{abstract}
\section{Introduction}
\label{sec:intro}
Intelligent systems in the real world often use machine learning (ML) algorithms to process various types of data.
However, graph data present a unique challenge due to their complexity. 
Graphs are powerful data representations that can naturally describe many real-world scenarios where the focus is on the connections among numerous entities, such as social networks, knowledge graphs, drug-protein interactions, traffic and communication networks, and more~\cite{ml_on_graphs}.
Unlike text, audio, and images, graphs are embedded in an irregular domain, which makes some essential operations of existing ML algorithms inapplicable~\cite{graphml_book}.
GML applications seek to make predictions, or discover new patterns, using graph-structured data as feature information: for example, one might wish to classify the role of a protein in a biological interaction graph, predict the role of a person in a collaboration network, or recommend new friends in a social network.

Unfortunately the majority of GML models are black boxes, thanks to their fully subsymbolic internal knowledge representation - which makes it hard for humans to understand the reasoning behind the model's decision process.
This widely recognized fundamental flaw has multiple negative implications: 
$(i)$ difficulty of adoption from domain experts~\cite{trust}, 
$(ii)$ non-compliance to regulation, (e.g. GDPR)~\cite{gdpr}, 
$(iii)$ inability to detect learned spurious correlations~\cite{wrong_reasons}, 
and $(iv)$ risk of deploying biased models~\cite{hcbias}. The eXplainable Artificial Intelligence (XAI) research field tackles the problem of making modern ML models more human-understandable.
The goal of XAI techniques is to extract from the trained ML model comprehensible information about their decision process.
Explainability is typically performed {\em a posteriori} - it is a process that takes place after the ML model has been trained, and possibly even deployed.
Despite a growing number of techniques for explaining GML models, most of them target node and graph classification tasks~\cite{yuan2022explainability}.
Link Prediction (LP) is a paradigmatic problem, but it has been relatively overlooked from the explainability perspective - especially since it has been often ascribed to knowledge graphs.
There are many ML techniques to tackle the LP problem, but the most popular approaches are based on an encoder/decoder architecture that learns node embeddings.
In this case, LP explanations are based on the interaction of pairs of node representations. 
It is still not clear how different graph ML architectures affect the explainer's behavior, and in the particular case of link prediction we have observed how explanations can be susceptible to technical choices for the implementation of both the encoding and the decoding stages.

Regarding the validation of explanation and explainers, few works have considered the study and evaluation of GML explainers for LP \cite{kang2021explanations}. Furthermore, despite growing interest regarding the validation of explanations, there is currently no consensus on the adoption of any standard protocol or set of metrics.
Given a formal definition for the problem of explaining link predictions, our Research Questions are therefore the following:
\begin{itemize}
    \item {\bf RQ1} How can we validate LP explainers and measure the quality of their explanations?
    \item {\bf RQ2} What hidden characteristics of LP models can be revealed by the explainers? What can we learn about the different LP architectures, given the explanations to their decisions?
\end{itemize}

In this paper, we propose a theoretical framing and a set of experiments for the attribution of GML models on the LP task, considering two types of Graph Neural Networks: Variational Graph Auto-Encoders (VGAE)~\cite{kipf2016variational} and Graph Isomorphism Networks (GIN)~\cite{gin_gnn}. 
We first perform a validation of the explanation methods on synthetic datasets such as Stochastic Block Models and Watts-Strogatz graphs, where we can define the ground truth for the explanations and thus compute the confusion matrices and report sensitivity (TPR) and specificity (TNR) for the attribution results. 
For real-world datasets with no ground-truth
(CORA, PubMed and DDI)~\cite{sen2008collective,wishart2017drugbank}, we exploit an adaptation of the insertion/deletion curves, a technique originally designed to validate computer vision models~\cite{petsiuk2018rise} that allows to quantitatively compare the produced explanations against a random baseline by inserting/removing features and/or edges based on their importance with respect to the considered attribution method.
\section{Related Work}
\label{sec:relwork}
\subsection{State-of-the-art Explainers for GML models}
\label{subsec:sota}
Considering the blooming research in the field of XAI for GNNs, and the increasing quantity of new methods that are proposed, we refer to the taxonomy identified in Yuan et~al. \cite{yuan2022explainability} to pinpoint the basic foundational principles underlying the different methods and choose few well-known models as representatives for broader classes of methods and use them in the remainder of the paper. Namely, we consider attribution methods based on perturbation methods, gradient-based approaches, decomposition - plus a hybrid one. A more detailed description of these classes and the selected explainers will be given in \ref{sec:explaining_link_pred}. 
We note that all these methods were originally discussed only in the context of node/graph classification.

Perturbation-based explainers study the output variations of a ML model with respect to different input perturbations. Intuitively, when important
input information is retained, the predictions should be similar to the original predictions. Existing methods for computer vision learn to generate a mask to select important input pixels to explain deep image models. Brought to GML, perturbation-based explainers learn masks that assign an importance to edges and/or features of the graph~\cite{luo2020parameterized,schlichtkrull2020interpreting,yuan2021explainability}. Arguably, the most widely-known perturbation-based explainer for GNNs is \emph{GNNExplainer}~\cite{ying2019gnnexplainer}.
Gradients/features-based methods decompose and approximate the input importance considering the gradients or hidden feature map values~\cite{integrated_gradients,smilkov2017smoothgrad,zhou2016learning,selvaraju2017grad,simonyan2013deep}. 
While other techniques just need to query the ML black-box at will ({\em model-agnostic} methods), explainers of this class require access to the internal weights of the ML model, and are therefore labelled as {\em model-aware} or {\em model-dependent}.
Another popular way to explain ML models is decomposition methods, which measure the importance of input features by decomposing the final output of the model layer-by-layer according to layer-specific rules, up to the input layer.
The results are regarded as the importance scores of the corresponding input features. 
\subsection{Link Prediction}
\label{subsec:lp}
Link prediction is a key problem for network-structured data, with the the goal of inferring missing relationships between entities or predict their future appearance.
Like node and graph classification, LP models can exploit both node features and the structure of the network; typically, the model output is an estimated probability, for a non-existing link.
Due to the wide range of real-world domain that can be modelled with graph-based data, LP can be applied to solve a high number of tasks. In social networks, LP can be used to infer social interactions or to suggest possible friends to the users~\cite{lp_social}. In the field of network biology and network medicine, LP can be leverage in predicting results from drug–drug, drug–disease, and protein–protein interactions to advance the speed of drug discovery~\cite{lp_drug}. 
As a ML task, LP has been widely studied~\cite{lp_survey}, and there exists a wide range of link prediction techniques. These approaches span from information-theoretic to clustering-based and learning-based; deep learning models represent the most recent techniques~\cite{lp_deep}.
The idea of enriching link prediction models with semantically meaningful auxiliary information has been seldom explored for simpler models, such as recommender systems~\cite{barbieri2014follow}, or with hand-crafted feature extraction~\cite{engelen2016explainable}. 
These approaches do not pair with the complex nature of deep GML models, where the feature extraction phase is part of the learning process, and models learn non-interpretable embeddings for each node.
Finally, even though there are many LP approaches more advanced than VGAE and GIN, these two architectures are the base of many popular LP approaches~\cite{ahn2021variational} and should be sufficient for the evaluation of the selected explanation techniques.

Regarding methods explicitly proposed to explain/interpret GNN-based LP models, Wang et~al. \cite{wang2021modeling} follow the intuition of focusing on the embeddings of pairs of nodes. Their explanations correspond to the attention scores of the aggregation step for the contexts interactions, and therefore they only give a first useful indication of important edges (and not features) for the prediction, but this preliminary information should be paired with a downstream explainer, as the authors point out.
For Xie et~al. \cite{xie2022task}, an explanation is a subgraph of the original graph that focuses on important edges, but ignores node features in the explanation, which are an important aspect in the decision process of a GNN. While the overall settings have differences, our work and their approach share the idea of considering embedding representations to produce graph explanations.

The task of explaining LP black-box models has been considered in the context of Knowledge Graphs \cite{rossi2022explaining,halliwell2022simplified}, but KGs consider labeled relations that must be taken into account and contribute actively to the explanation. When considering unlabeled edges, a different approach for explaining the LP task is required.
Regarding the LP frameworks that incorporate features such as distance encoding and hyperbolic encoding~\cite{zhang2018link,chami2019hyperbolic,yan2021link,zhu2021neural}, we believe that there should be a community-wide discussion about how such features can be incorporated in the proposed explanations. In our view, while these frameworks are very powerful for capturing features that are important for the LP task, none of the current attribution methods is able to assign an explanation to such features.

Closely related to our work, recent attention has been posed onto the topic of systematically evaluating the produced explanations~\cite{sanchez2020evaluating,faber2021comparing,agarwal2022probing,agarwal2023evaluating}, but exclusively for node/graph classification tasks. Here we fill the gap for the LP task.
\section{Explaining Link Predictions}
\label{sec:explaining_link_pred}
Given a graph $G = (V,E)$ with set of nodes $V$ and set of edges $E \subseteq 2^{V \times V}$, and a node-feature matrix $X \in \mathbb{R}^{|V| \times F}$, link prediction estimates the probability that an unseen edge between two nodes $i, j \in V$, $(i,j) \notin E$ is missing (e.g. when reconstructing a graph or predicting a future edge).
Formally, a link prediction model is a function $\phi_{G,X}: V \times V \mapsto [0,1]$ that given $G$ and $X$ maps every pair of nodes in $V$ to a probability $p \in [0,1]$.
A common approach for LP tasks is to learn a node representation in a vector space (\textit{encoder} $\text{Enc}_{G,X}: V \mapsto \mathbb{R}^d$), and then estimate edge probability from pairwise distances in this latent space (\textit{decoder} $\text{Dec}: \mathbb{R}^d \times \mathbb{R}^d \mapsto  [0,1]$). Most encoders are currently based on a message-passing mechanism that learns a latent representation of the nodes via a aggregate-update iterative mechanism. At each iteration, of each node in the graph receives messages coming from neighboring nodes (their current embeddings). The messages are then aggregated through a permutation-invariant function and the node embedding is subsequently updated using a non-linear learnable function of the current node embedding and the aggregated messages, such as a multi-layer perceptron \cite{zhou2020graph,bronstein2021geometric}. Decoders for link prediction usually compute a similarity function between node embeddings, such as the inner product between two node embeddings, followed by a normalization function, such as a sigmoid function, to obtain the probability of a link between the two nodes.
\subsection{Attribution methods for link prediction}
\label{subsec:extending_explainers}
A LP explainer implements a function that, given an edge $(i,j)$ and a model to explain, maps the edges in $E$ and the node features in $X$ to their respective explanation scores. The higher the explanation score, the more important the edge (or the feature) is for the model to estimate the probability of $(i,j)$.
For this work, we have selected representative LP explainers basing our choice on {\em (i)} their belonging to different classes of the taxonomy described by Yuan et~al. \cite{yuan2022explainability} to have a representative set of explainers, and {\em (ii)} their adoption and availability of code. Namely, we consider attribution methods based on perturbation (\textit{GNNExplainer}~\cite{lucic2022cf-gnnexplainer}), gradient-based approaches (\textit{Integrated Gradients} (IG)~\cite{sundararajan2017axiomatic}), decomposition (\textit{Deconvolution} \cite{zeiler2014visualizing}) - plus a hybrid one (\textit{Layer-wise relevance propagation} (LRP) \cite{bach2015pixel}).

\textit{GNNExplainer}~\cite{lucic2022cf-gnnexplainer} searches for a subgraph $G_S$, and a subset of features $X_S$ of the original dataset $G$, $X$ that maximises the mutual information between the outputs of $G, X$ and $G_S, X_S$.
Since LP outputs a probability, the goal is reduced to finding a $G_S$ that maximizes the probability of the model output while enforcing sparseness in $G_S$.
Therefore, explanation scores are defined as a mask on edges and node features.
GNNExplainer provides explanations for LP with no change to its optimization goal, but the model's encoder and decoder must be plugged in so that the edge and feature masks can be properly estimated. In our setting, when a model predicts a link $(i,j)$, GNNExplainer learns a single mask over all links and features that are in the computation graphs of $i$ and $j$.

\textit{Integrated Gradients} (IG)~\cite{sundararajan2017axiomatic} is an axiomatic attribution method that aims to explain the relationship between a model's predictions in terms of its features. It oughts to satisfy two axioms: sensitivity and implementation invariance, by analysing the gradients of the model with respect to its input features.
In the case of link prediction, IG assigns positive and negative explanation scores to each link and each node feature, depending on how sensible the model's prediction is as these inputs change.

\textit{Deconvolution} \cite{zeiler2014visualizing}, first introduced for the explanation of convolutional neural networks in image classification, is a saliency method that uses a deconvolution operation to perform a backward propagation of the original model. It allows to highlight which feature or edge is activated the most and the attribution output consists in positive and negative scores for edges and node features.

\textit{Layer-wise relevance propagation} (LRP) \cite{bach2015pixel} is based on a backward propagation mechanism applied sequentially to all layers of the model. For a target neuron, its score is represented as a linear approximation of neuron scores from the previous layer. Here, the model output score represents the initial relevance which is decomposed into values for each neuron of the underlying layers, based on predefined rules. In this paper we use the $\varepsilon-stabilized$ rule as in \cite{baldassarre2019explainability}.

\begin{figure}[!ht]
     \centering
      \includegraphics[width=.99\textwidth]{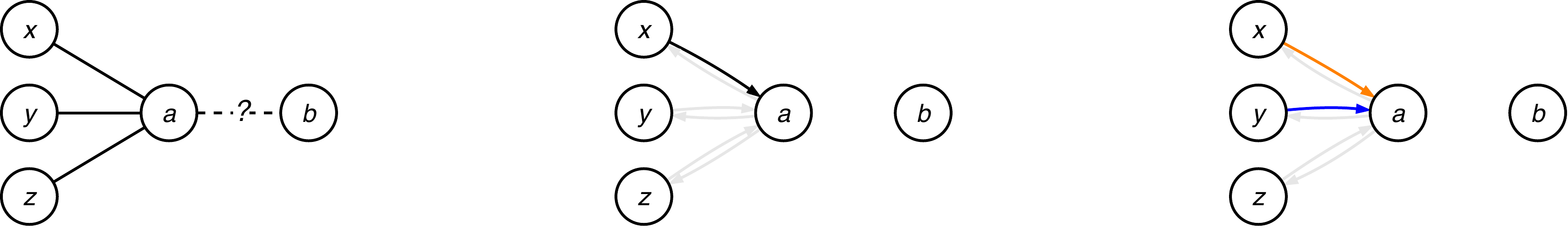}
    \caption{Toy graph (left) and explanations (mask and attribution) for the link $(a, b)$ in the toy graph for GNNExplainer (center) and Integrated Gradients  (right) attribution methods. The edge color in the right panel indicates positive (orange) and negative (blue) importance.
    The explanations produced by Deconvolution and LRP are similar to Integrated Gradients.
    }
    \label{fig:toy_model}
\end{figure}

To illustrate how the above attribution methods work in practice for LP, we start with a white-box message-passing model for link prediction on a toy example given by the graph shown in Figure~\ref{fig:toy_model}(left) with 5 nodes and 3 edges, $V=(a, b, x, y, z)$, $E=\{(a, x), (a, y), (a, z)\}$ and a feature matrix $X$ with two node features defined as
\begin{equation}
    X=\begin{bmatrix} 0.5 & 0.5 \\ 1 & 0 \\ 1 & 0 \\ 0 & 1 \\ 0.5 & 0.5\end{bmatrix}.
\end{equation} 
We define the embeddings $e_i \in \mathbb{R}^d, i\in V$ of the graph nodes as 
\begin{equation}
    e_{ia} = \frac{1}{|\partial i|}\sum_{j\in \partial i}X_{ja} + X_{ia}, \ a=1, \dots, d,
\end{equation}
where $\partial i \equiv \{j\in V: i\neq j, (i, j)\in E\}$ indicates the set of first neighbors to node $i$. The probability of an edge between two nodes (the decoder) is the cosine similarity between the node embeddings. 

We ask to explain the prediction for link $(a, b)$. Here the edge $(a, x)$ has positive score because it ``pulls" the embedding of $a$ closer to $b$, while the edge $(a, y)$ has negative score because it ``pushes" the embedding of $a$ away from $b$. The edge $(a, z)$ is neutral.
Figure~\ref{fig:toy_model} shows the edge explanations provided by GNNExplainer (center) and IG (right). IG is able to reflect the ground truth as it provides both positive and negative scores, while GNNExplainer considers positive masks only, thus returning a partial result. The explanations from Deconvolution and LRP are similar to the one produced with IG.

\subsection{Validating Explanations}
\label{subsec:validating}
The validation of explanations is a generally overlooked topic in XAI, and LP tasks are no exceptions. Here, we suggest two different approaches, respectively to deal with ground-truth cases, and with no-ground-truth cases.

When ground truth is available, we use metrics from information retrieval. The ground truth is defined as a binary mask over $E$ and $X$ where $(i,j)$ is true if the edge is important to the model prediction (and false otherwise), and a binary mask over features that follows the same logic.
The explanation scores are binarized, fixing a standard threshold for the explanation scores, e.g. $0.5$ for the positive defined masks of GNNExplainer and $0$ for the other explainers considered here, or selecting the optimal threshold based on the ROC curve of true positive rate and false positive rate obtained varying the threshold, so that we can calculate a confusion matrix. True positives are considered when a high explanation score is assigned to an edge (or a feature) that is important according to the ground truth. False positives are considered when high explanation scores are assigned to non-important edges (or features). True negatives (and accordingly, false negatives) are considered when low explanation scores are assigned to unimportant (or important) edges or features.
Finally, metrics such as precision, recall, specificity and sensitivity are calculated for each explainability technique.
Here we focus on specificity and sensitivity, i.e., the true positive and true negative rates.

When ground truth explanations are not available, we resort to a validation method borrowed from explainability for computer vision, proposed first by Petsiuk et~al. \cite{petsiuk2018rise}, that we adapt for graph explanations. To the best of our knowledge this is the first time this validation method is used in this context. This method consists in progressively removing/inserting features and/or edges based on their importance with respect to the attribution method considered. The feature and edge attributions are sorted by decreasing score and in the \emph{deletion} case they are gradually removed. In the \emph{insertion} case, they are gradually inserted in decreasing order of score starting with no features/edges. Intuitively, if the explainer's output is correct, removing or adding the most important features will cause the greatest change in the model output. The area under the curve of the fraction of features inserted/removed versus the output of the model provides a quantitative evaluation of the explanation.

\begin{figure}[!ht]
     \centering
     \hspace*{\fill}
     \begin{subfigure}[b]{0.49\textwidth}
         \centering
         \includegraphics[width=\textwidth]{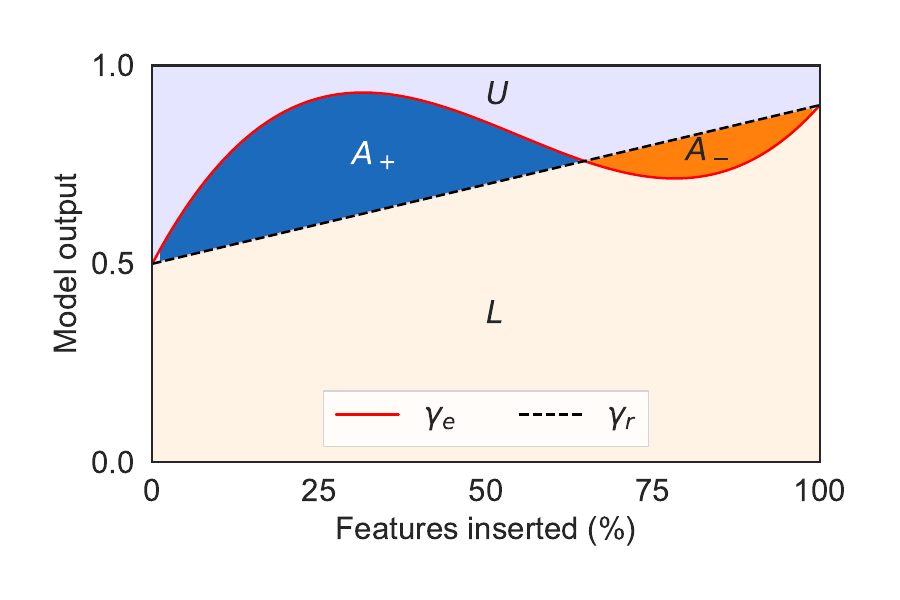}
     \end{subfigure}
     \hfill
     \begin{subfigure}[b]{0.49\textwidth}
         \centering
         \includegraphics[width=\textwidth]{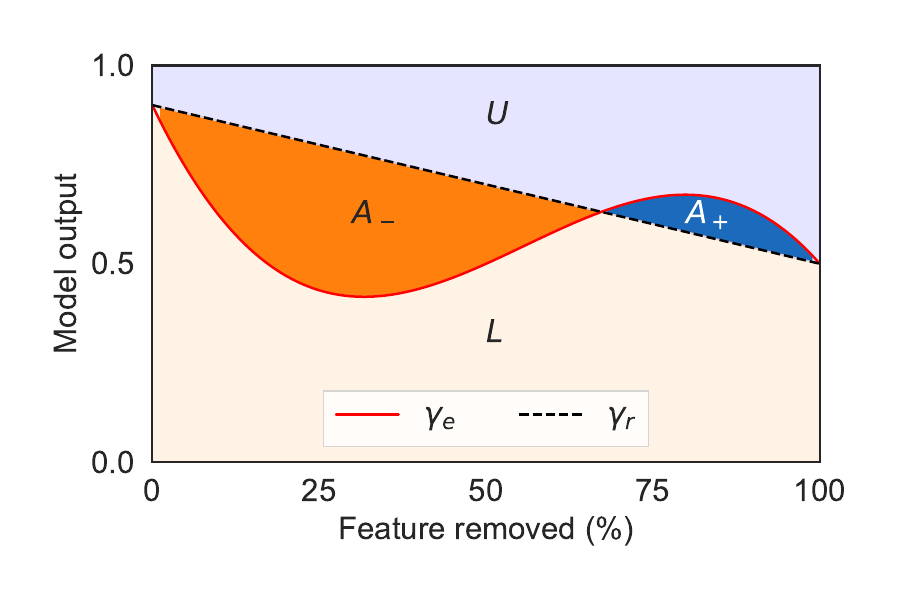}
     \end{subfigure}
     \hspace*{\fill}
   \caption{Illustration of the area score for the feature insertion (left) and feature deletion (right) procedures.
   $\gamma_e$ is the insertion (deletion) curve when features are sorted according to their explanation scores, while $\gamma_r$ is the random insertion (deletion) curve.   
   The area score for the insertion procedure is given by $\frac{A_+}{U} - \frac{A_-}{L}$.
   The area score for the deletion procedure is $\frac{A_-}{L} - \frac{A_+}{U}$.
   }
    \label{fig:area_score}
    \vspace{-0.2cm}
\end{figure}

To quantitatively compare different attribution methods,
we define the following \emph{area score}: referring to Figure~\ref{fig:area_score}, for the insertion case, consider the area $A_+$ comprised between the explainer curve $\gamma_e$ and the random curve $\gamma_r$ when $\gamma_e>\gamma_r$, and the area above $\gamma_r$, U. The ratio $\frac{A_+}{U}\in[0, 1]$ describes the portion of the graph where the explainer performs better than the random baseline. Consider then the area $A_-$ comprised between the explainer curve $\gamma_e$ and the random curve $\gamma_r$ when $\gamma_e<\gamma_r$, and the area below $\gamma_r$, L. The ratio $\frac{A_-}{L}\in[0, 1]$ corresponds to the portion of the graph where the explainer performs worse than the random baseline. We define the final score as 
\begin{equation}
    s_{ins}\equiv \frac{A_+}{U} - \frac{A_-}{L} \in [-1, 1].
\end{equation} 
Similarly, for the deletion case the score is given by 
\begin{equation}
    s_{del}\equiv \frac{A_-}{L}-\frac{A_+}{U} \in [-1, 1]. 
\end{equation}

The area score is a summary metric for the insertion and deletion procedures, and reflects the ability of the explainer to assign higher scores to the most influential features/edges for the considered prediction. Ideally, a perfect explainer should give high scores to very few edges and/or features that carry almost all the information necessary for the prediction. In this case, inserting these features would be sufficient to recover the output of the model when all the features/edges are present, and deleting it would cause a great drop in the output of the model. In this case the area score would be equal or close to 1. In the case of random explanation scores, removing/inserting the features/edges with the highest score would not have, on average, a strong impact on the output of the model. In this case the area score would be 0. A negative value of the area score indicates a performance worse than the random baseline.

Note that the absolute values of the area score for the insertion and deletion procedure are not directly comparable since the normalization is different. This score is particularly useful for comparing the performance of different explainers with respect to the random baseline under the same procedure. The deletion curve is closely related to the fidelity and sparsity metrics \cite{yuan2022explainability}, but the area score has the advantage of providing a single metric that coherently summarizes the two for easier readability. The insertion curve complement the deletion curve, in the sense that instead of considering the distance between the original model output and the output obtained by iteratively removing the most important feature by explanation score, it considers the distance between the original model and the output obtained by starting with all null features and iteratively adding the most important features by explanation score.
\section{Experiments}
In this section we report the results of evaluating LP explanations in two distinct scenarios -- one with ground-truth explanations, and another without. In the first scenario, we use synthetic data, where graph datasets are generated along with their respective ground truth explanations for the created edges. This approach allows us to assess the explanations in a controlled setting, where we know the true explanations.

In the second scenario, we turn to empirical data from three different datasets. 
Here, without the availability of ground-truth explanations, we assess the quality of explanations produced by the explanation methods through the area score defined in Section~\ref{subsec:validating}. This provides a means to measure the performance of explanation methods in real-world, less controlled conditions.

\begin{figure}[!ht]
     \centering
     \begin{subfigure}[b]{0.45\textwidth}
         \centering
         \includegraphics[width=\textwidth]{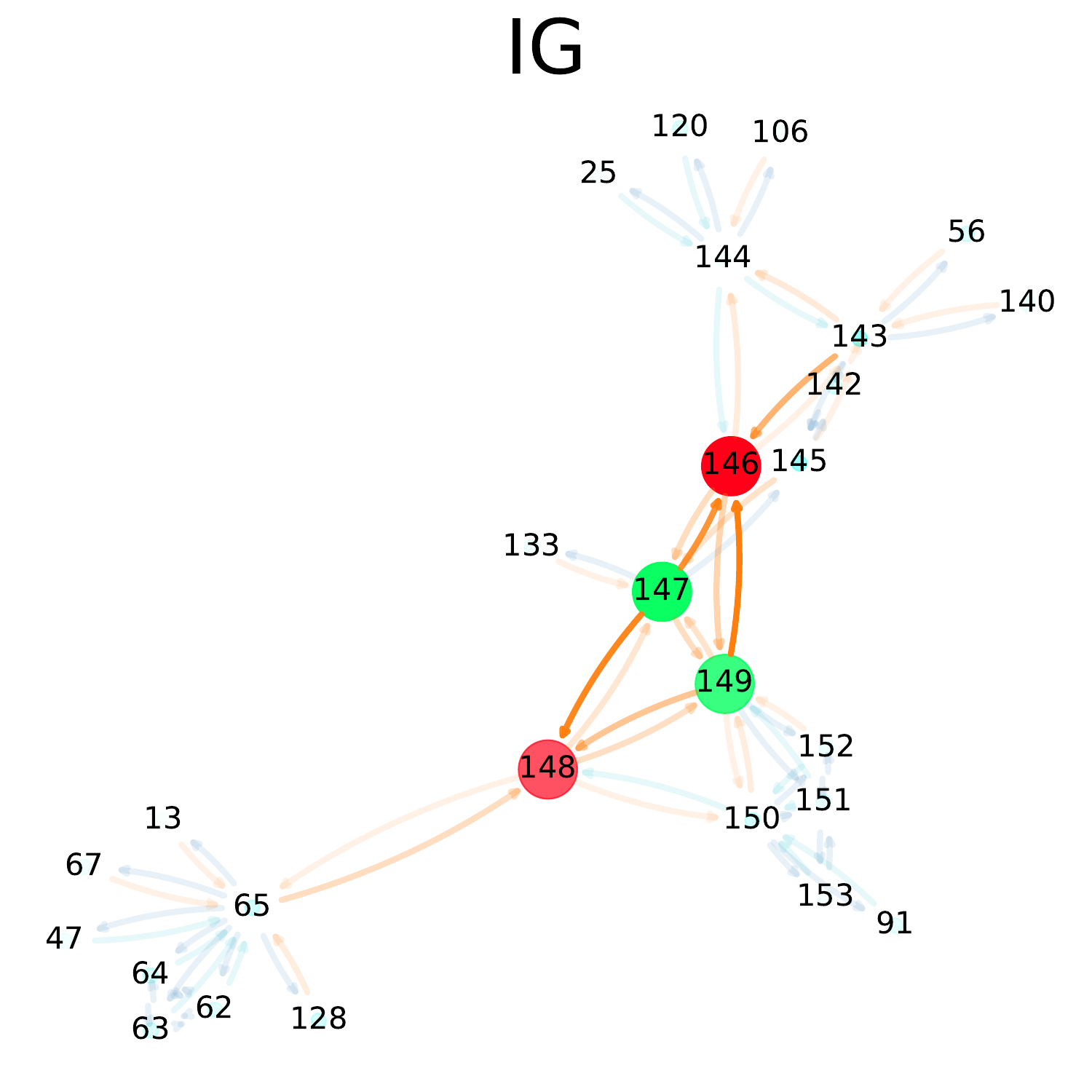}
     \end{subfigure}
     \begin{subfigure}[b]{0.45\textwidth}
         \centering
        \includegraphics[width=\textwidth]{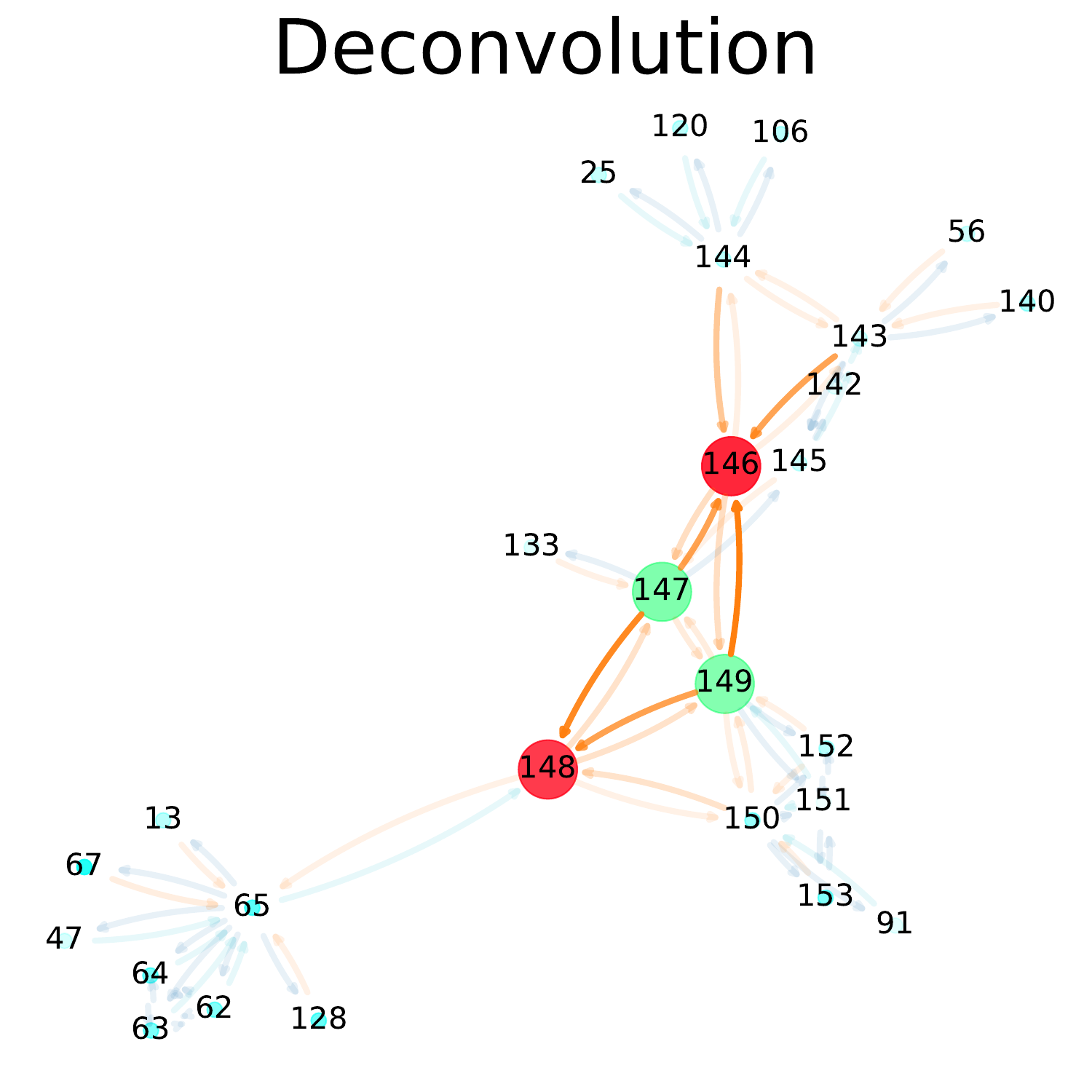}
     \end{subfigure}\\
     \begin{subfigure}[b]{0.45\textwidth}
         \centering
         \includegraphics[width=\textwidth]{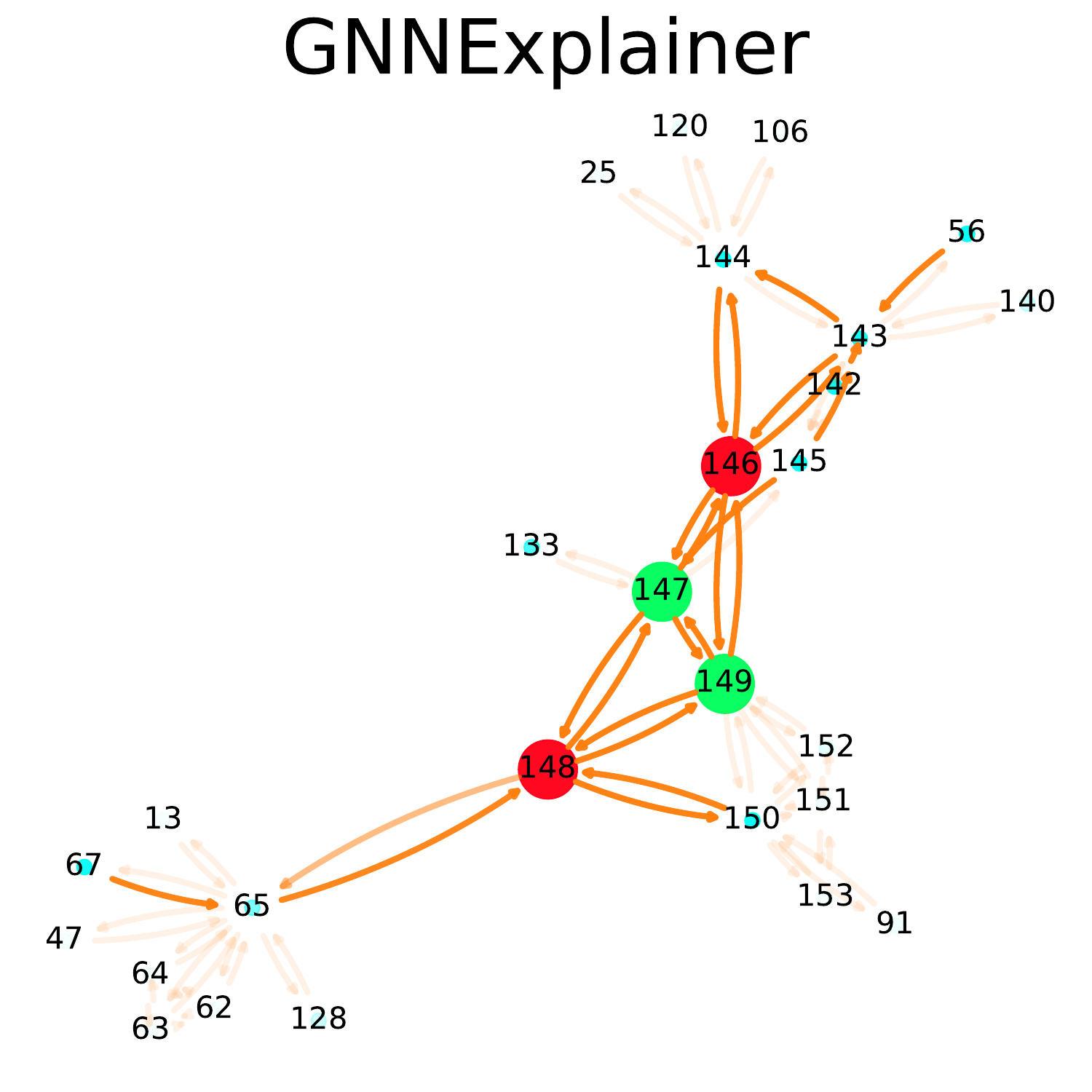}
     \end{subfigure}
     \begin{subfigure}[b]{0.45\textwidth}
         \centering
        \includegraphics[width=\textwidth]{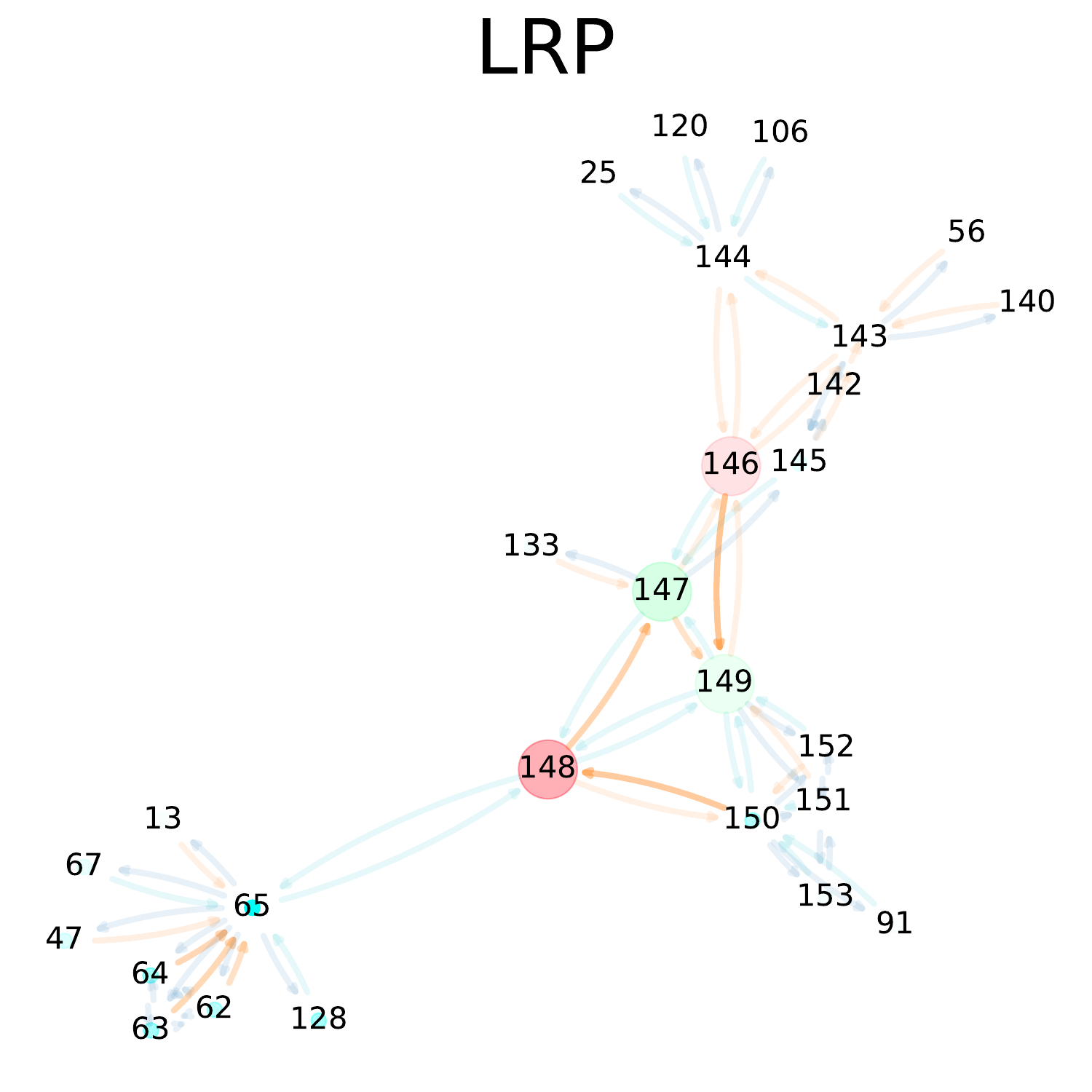}
     \end{subfigure}
    \caption{Explanations for GIN trained on Watts-Strogatz graph data for the presence of an edge between the two nodes in red. The two nodes should be connected based on the triangle closure (green nodes). IG and Deconvolution give good explanations, while GNNExplainer considers important most of the computational graph, and LRP fails to identify the important edges.}
    \label{fig:single_explanation_example}
\end{figure}

Our experiments consist of four steps: (i) dataset preparation, (ii) model training, (iii) attribution, and (iv) attribution evaluation. Edges are split into training and test sets, with the same proportion of positive and negative edges, and attributions are performed on the test set. To ensure reproducibility and fair comparison, we test all explainers with the same trained model and train-test sets for each dataset. Multiple realizations of the attribution process with different random seeds account for the stochastic nature of ML training. For each dataset, we consider 2 encoders (VGAE, GIN), 2 decoders (Inner product, Cosine distance), and 4 explainers (GNNExplainer, Integrated Gradients, Deconvolution, and LRP). In Fig. \ref{fig:single_explanation_example} we show an example of the explanations given by each of the explainers considered for a GIN network predicting a missing edge on the Watts-Strogatz dataset (see section \ref{sec:synth}).

\subsection{Synthetic Data}
\label{sec:synth}
We consider two generative models, namely the Stochastic Block Model (SBM)~\cite{holland1983stochastic} and the Watts-Strogatz model (WS)~\cite{watts1998collective}, as examples of graphs where we can reconstruct the ground truth attributions for the link prediction task. In these experiments, whether an edge is present or not is clearly defined by the generative model. The small proportion of random edges introduced by the two stochastic generative models are not used to evaluate the explainers.
We assume that a model trained on a sufficient number of data points is able to reflect the logic of the generative model, therefore an explainer should reflect this aspect in its attributions. For the SBM, a link should be present if two nodes belong to the same block, while for WS a link should be predicted if two nodes belong to a triangle completion.  The node features in both cases are simply the one-hot encodings of the node ids, i.e. $X$ corresponds to the identity matrix.
This is a common choice to use as node features in the absence of meaningful ones \cite{you2021identity}.

For both models, the experiments are designed as follows: we generate a graph $G=(V, E)$ of given size $|V|$, and we train a GML model for the LP task on a training fold of $G$. Then, the explainer is asked to explain each edge in the test set (except for the random edges). We compare the attributions to the ground truth, computing a confusion matrix of the results. We get a score for each predicted edge, obtaining the error distribution for the explainer.

\begin{figure}[!ht]
     \centering
     \begin{subfigure}[b]{0.8\textwidth}
         \centering
         \includegraphics[width=\textwidth]{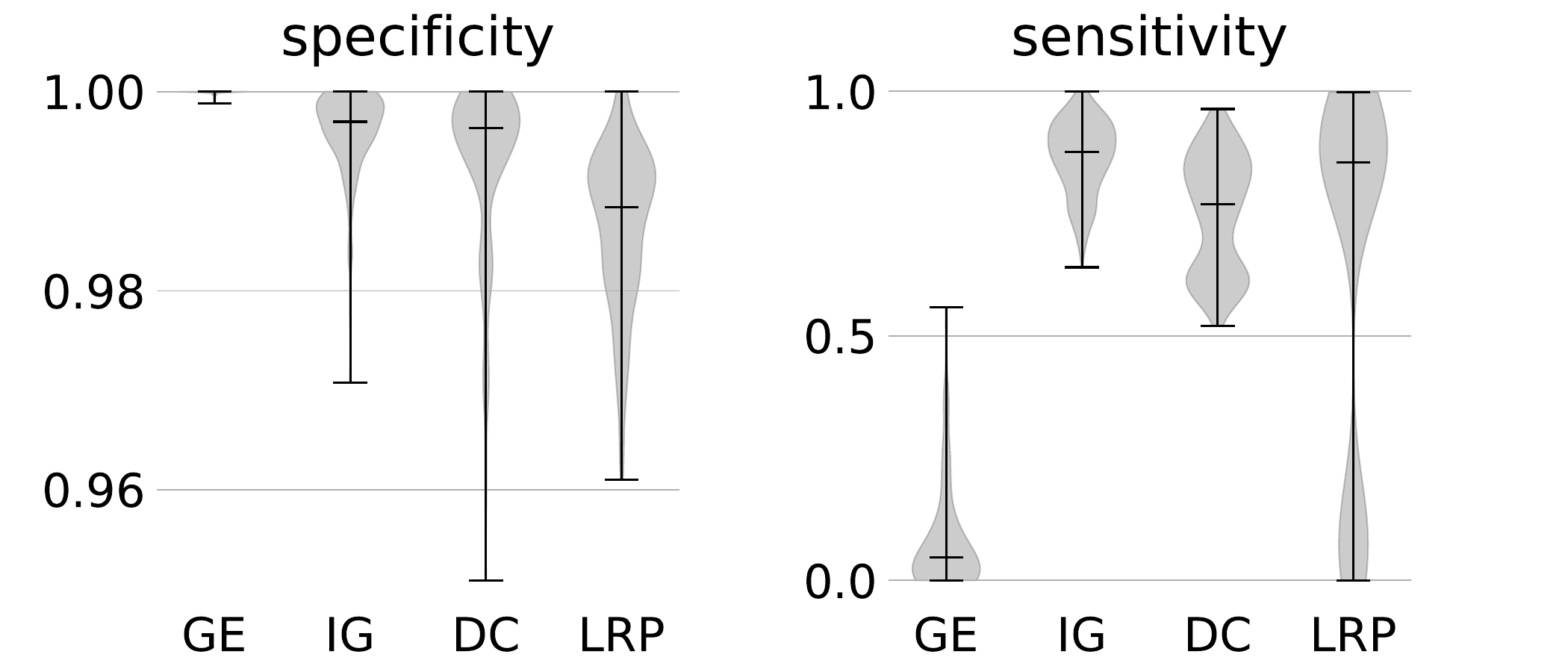}
     \end{subfigure}\\
     \begin{subfigure}[b]{0.8\textwidth}
         \centering
        \includegraphics[width=\textwidth]{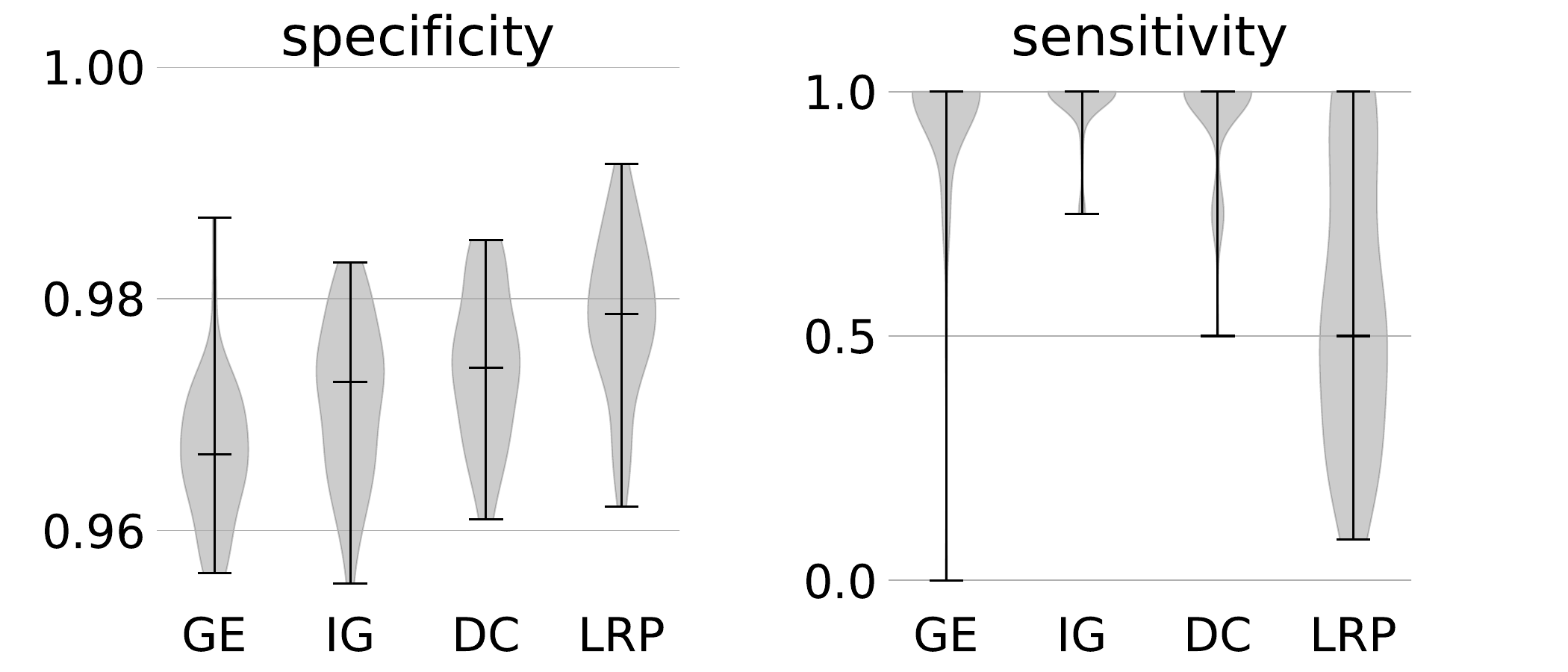}
     \end{subfigure}
    \caption{Sensitivity and specificity distributions on the Stochastic Block Model (top) and Watt-Strogatz (bottom) graphs for the four considered attribution methods: GNNExplainer (GE), Integrated Gradients (IG), Deconvolution (DC) and Layer-wise Relevance Propagation (LRP).}
    \label{fig:sensitivity_specificity}
\end{figure}

For the sake of readability, we summarize the results with two metrics, namely specificity and sensitivity. Specificity measures the proportion of true negatives, that is, the number of edges that receive small importance from the explainer and that are in fact not important for the considered edge, over the number of true negatives; similarly, sensitivity is the ratio between predicted and true positives.
Sensitivity and specificity together completely describe the quality of the attribution, but should not be considered separately. 
Figure~\ref{fig:sensitivity_specificity} shows the sensitivity and specificity distributions for the four explainers tested on a GIN model trained on SBM (left) and WS (right) graphs.

In the SBM case (left), GNNExplainer demonstrates better specificity than other explainers but suffers from poor sensitivity due to numerous true positives in an SBM block. This is because it tends to produce sparse masks, often missing many true positives. This issue is particularly evident in SBM, where explanations involve numerous nodes, while in the WS graph, GNNExplainer's performance is in line with other explainers due to the sparse explanation.

\begin{figure}[!ht]
     \centering
      \includegraphics[width=0.8\textwidth]{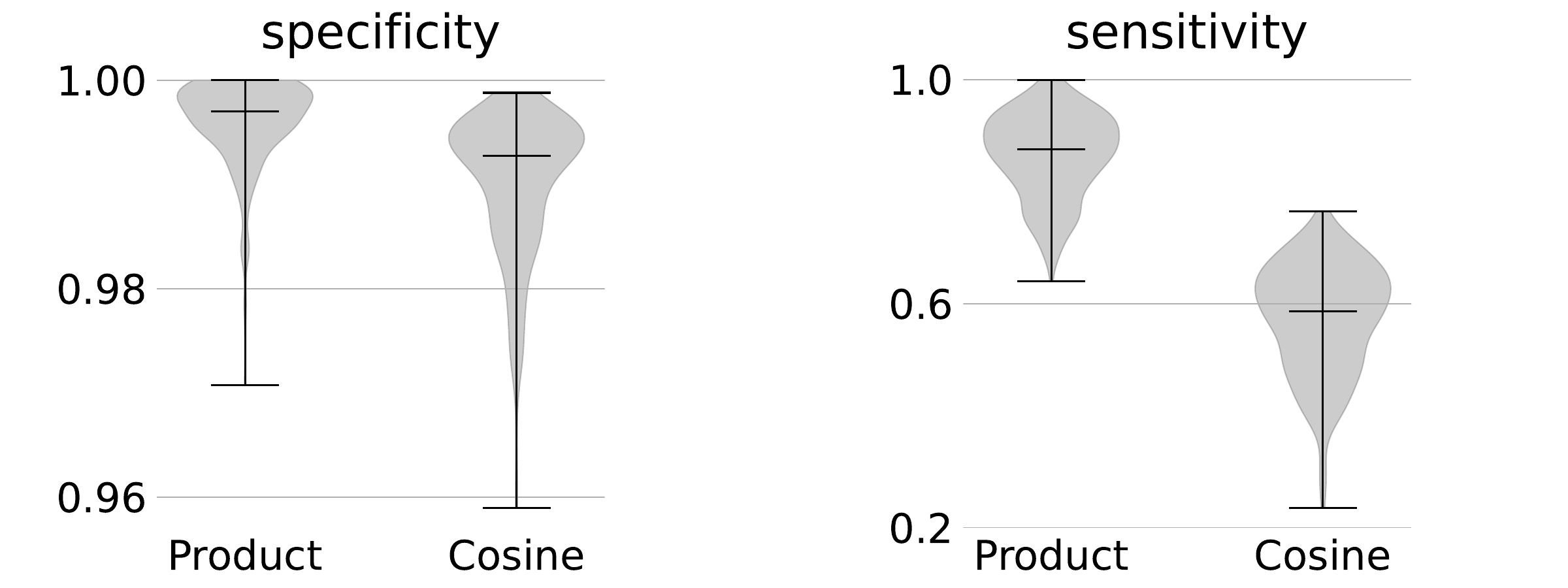}
    \caption{Difference in the performance (specificity and sensitivity) of the IG attribution method for two GCN models with same encoder but different decoders, one based on scalar product and other based on cosine similarity, respectively.}
    \label{fig:cos_vs_vecprod}
\end{figure}

The similarity measure used in the decoder significantly impacts the explanation quality. Common measures like cosine similarity pose challenges for current explainability techniques. Issues arise when nodes become more similar as information is masked, leading to degenerate solutions like empty subgraphs and the masking of all features. Consequently, for explainers that search for a subgraph that maximizes the model output such as GNNExplainer, no edges or features are deemed important when using cosine similarity between node embeddings. To highlight the impact of the decoder in producing explanations, we show in Figure~\ref{fig:cos_vs_vecprod} the sensitivity and specificity distributions for the IG explainer (but the results are similar for all explainers) when applied to two different GNN models that differ only in the decoder: the first uses a inner product of the node embeddings followed by a sigmoid, and the second uses a cosine similarity decoder.
The explanation quality drops drastically if the model uses the cosine distance. This metric, due to normalization, is prone to produce explanation scores that are close to zero.

\subsection{Empirical Data}
\label{sec:empirical}
In this section we focus on the validation of explainability methods when ground truth explanations are not available. In order to do so, we consider three empirical datasets: Cora and PubMed~\cite{sen2008collective}, plus a drug-drug interaction (DDI) network obtained from DrugBank \cite{wishart2017drugbank}.
The graph $G$ and the node features $X$ are constructed according to Yang et~al. \cite{yang2016revisiting}: the bag-of-words representation is converted to node feature vectors and the graph is based on the citation links. The Cora dataset has 2,708 scientific publications classified into seven classes, connected through 5,429 links. The PubMed dataset has 19,717 publications classified into three classes, connected through  44,338 links.
Although originally introduced for the node classification task, these two datasets are common benchmarks for the evaluation of current state-of-the-art GML models for the LP task, allowing a precise comparison of the performance of the models that we are explaining. The DDI dataset has 1,514 nodes representing drugs approved by the U.S. Food and Drug Administration, and 48,514 edges representing interaction between drugs. The dataset does not provide node features, that are provided as node embedding vectors of fixed dimension 128 computed using Node2Vec \cite{grover2016node2vec}.
For this dataset, link prediction is a critical task, aiming to anticipate potential drug-drug interactions that have yet to be observed. 
Predicting these interactions can mitigate their adverse effects and health risks, thereby promoting patient safety through preventive healthcare measures. 
\begin{figure}[!t]
     \centering
      \includegraphics[width=0.9\textwidth]{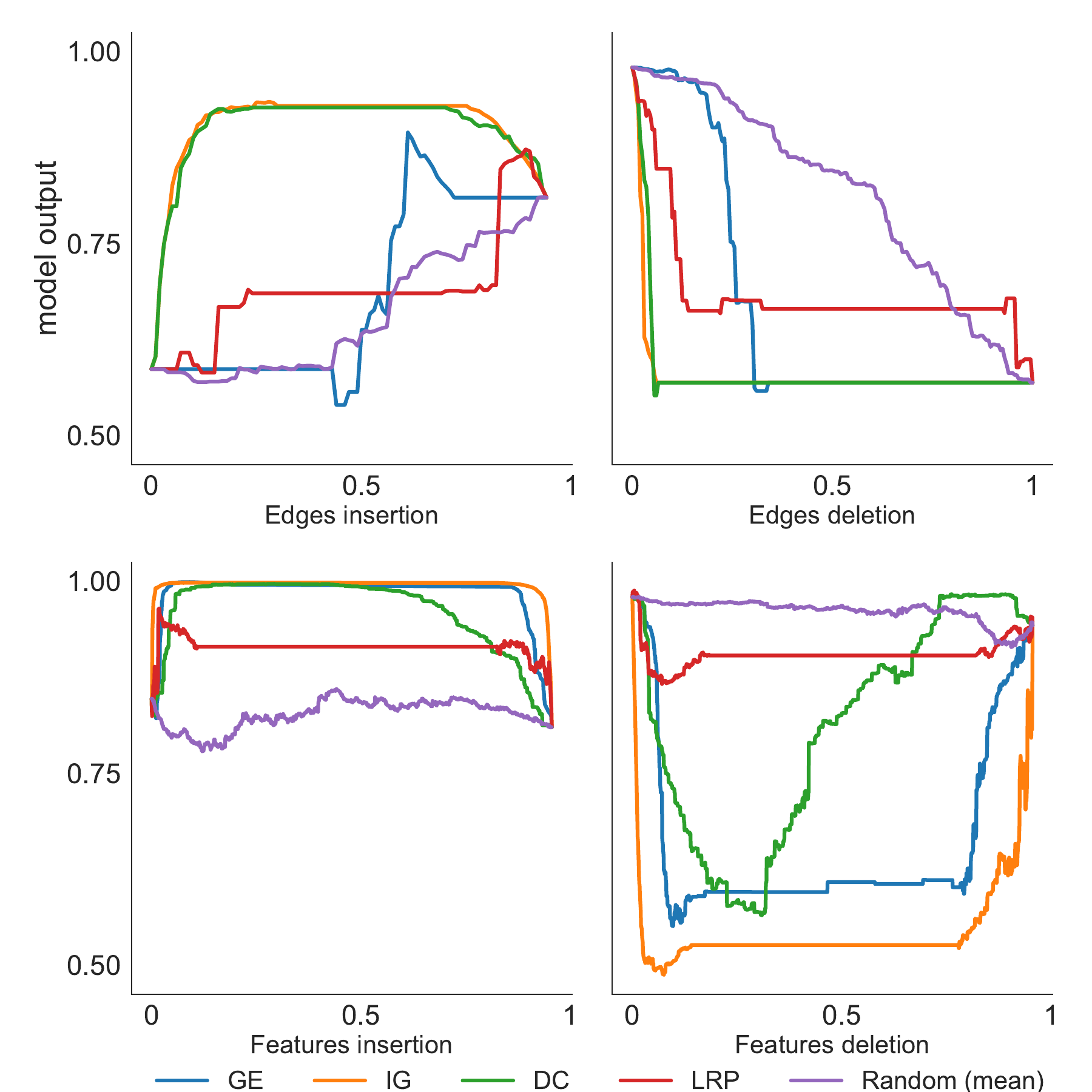}
    \caption{Examples of insertion and deletion curves for edges (top) and features (bottom) for a GIN model with Inner product decoder. 
    }
    \label{fig:insertion_deletion_example}
\end{figure}
\begin{figure}[!ht]
     \centering
     \begin{subfigure}[b]{0.8\textwidth}
         \centering
         \includegraphics[width=\textwidth]{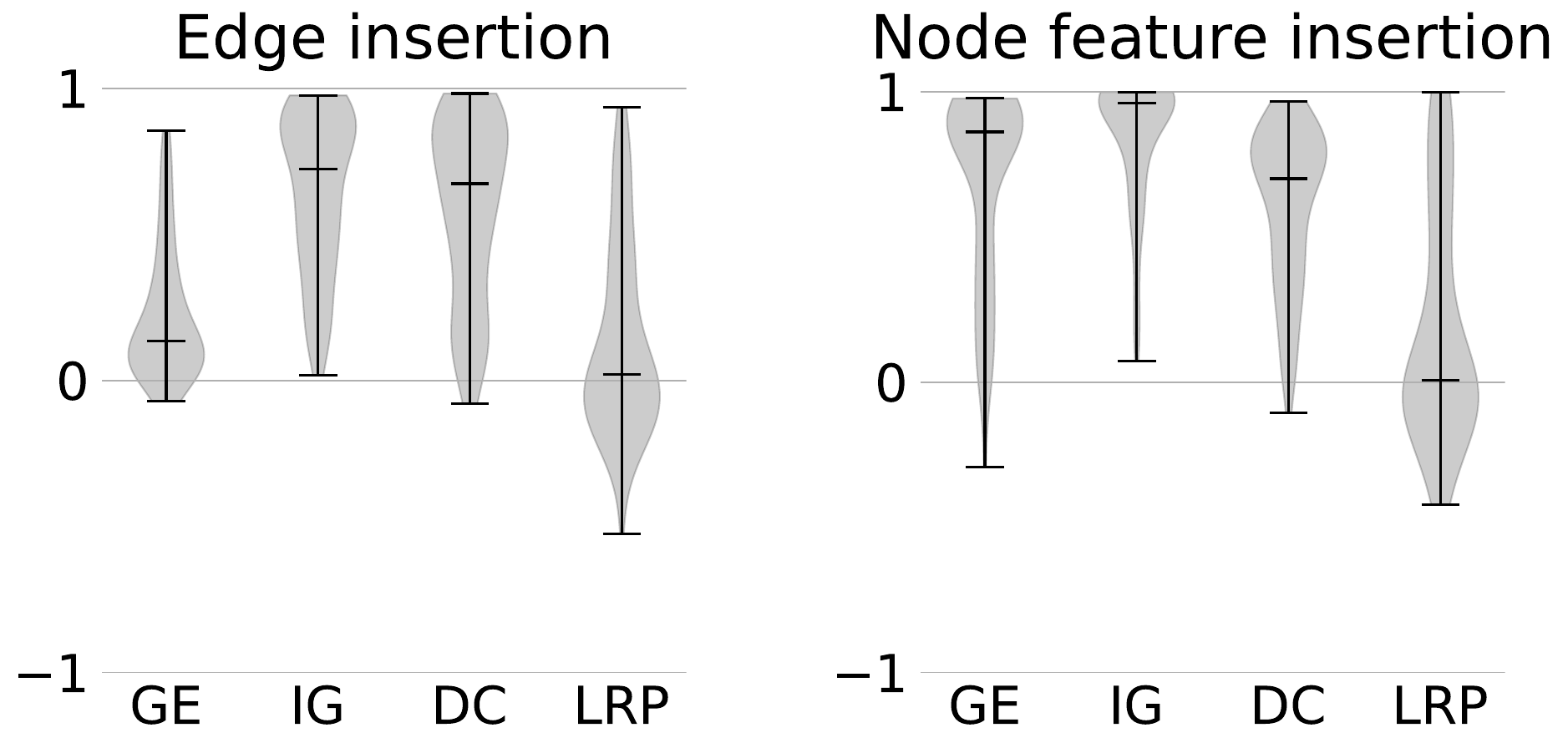}
     \end{subfigure}\\
     \begin{subfigure}[b]{0.8\textwidth}
         \centering
        \includegraphics[width=\textwidth]{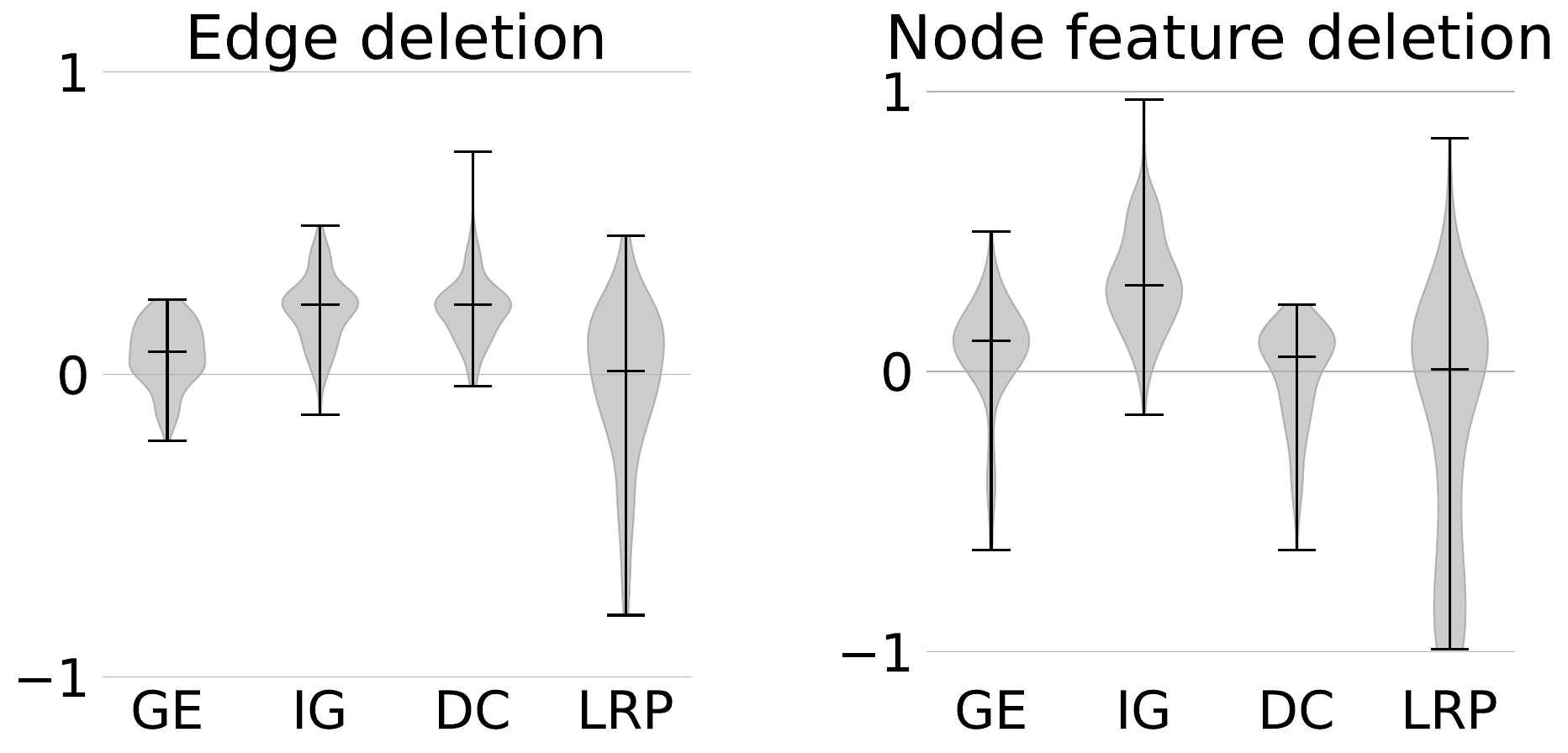}
     \end{subfigure}
    \caption{Score distribution for feature and edge insertion and deletion.
    For each edge in the test set, we produce their respective insertion and deletion curves for features and edges, and calculate the area score according to the procedure illustrated in Figure~\ref{fig:area_score}.
    The violin plots show the distribution of scores for each explainability method: GNNExplainer (GE), Integrated Gradients (IG), Deconvolution (DC) and Layer-wise Relevance Propagation (LRP).
    }
    \label{fig:area_score_insertion_deletion}
\end{figure}

We train two state-of-the art types of GNN encoders that are suitable for the link prediction task, namely the VGAE \cite{kipf2016variational} and GIN \cite{xu2018powerful}. For both encoder architectures, we use the inner product of node embeddings followed by a sigmoid as the decoder.
Training the GNNs on CORA we reach an AUC test score of $0.952$ ($accuracy = 0.744$) for VGAE and an AUC test score of $0.904$ ($accuracy = 0.781$) for GIN.
Training on PubMed we reach AUC test score of $0.923$ ($accuracy = 0.724$) for VGAE and AUC test score of $0.893$ ($accuracy = 0.742$) for GIN. Lastly, on DDI we reach AUC test score of $0.881$ ($accuracy = 0.667$) for VGAE and AUC test score of $0.920$ ($accuracy = 0.751$) for GIN.
Once the model is trained we consider the edges in the test set and look at the explanation scores for node features and edges resulting from the attribution methods. 
These scores define insertion and deletion curves as described in Section \ref{subsec:validating}. In Figure~\ref{fig:insertion_deletion_example} we show an example of insertion and deletion curves for node features and edges attributions obtained for a single edge predicted by a GIN model trained on CORA. The x axis refers to the ratio of edges/features that have been inserted/removed by attribution importance, and the y axis shows the variation in the model output, for the selected class, in presence/absence of these features or edges. Each curve represents a single explainer, plus a curve (in purple) that represents the random insertion/deletion baseline. The random baseline is computed by adding/removing features or edges at random without taking in consideration any attribution score. Many realization of the random curve are then averaged in order to obtain a robust baseline.
We then compute the area score defined in \ref{subsec:validating} for all the considered attribution methods and all the edges in the test set.
In Figure~\ref{fig:area_score_insertion_deletion} we show the distribution of scores obtained from the CORA dataset with a GIN model.
In Table \ref{tab:area_score_insertion_deletion} we report the results for all tested explainability methods for the three datasets using the GIN architecture as the encoder.

\begin{figure}[!ht]
     \centering
      \includegraphics[width=0.8\textwidth]{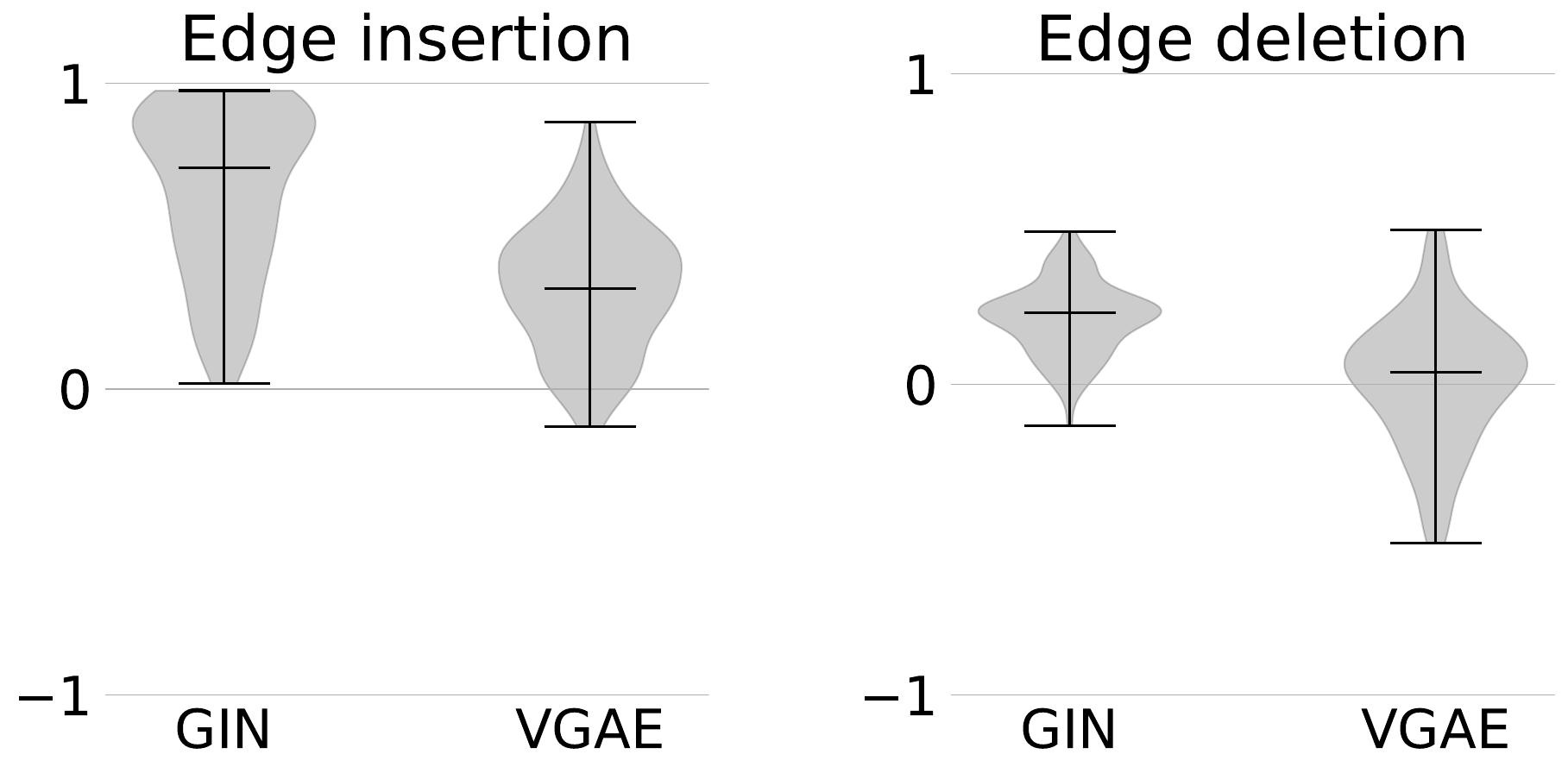}
    \caption{Area score difference between VGAE and GIN models for edge insertion (left) and deletion (right) on the CORA dataset. The VGAE model has almost no gain with respect to the random baseline, while the GIN model is consistently better. This is suggestive of a different exploitation of topological features, that is, edges, in the graph by different encoder architectures when learning the link prediction task.}
    \label{fig:area_score_vgae_gin}
\end{figure}

The case of edge deletion/insertion is particularly interesting when comparing the two different GNN architectures. Even if they perform comparably on the task of link prediction, the area score for the attribution on the VGAE model drops drastically for all attribution methods, suggesting that most of the signal of the data is taken from the node features alone, while the GIN model shows a very different scenario, where also the edges, and thus the network structure, are important for the model. In Figure~\ref{fig:area_score_vgae_gin} we show the distribution of gain in the area score when inserting/deleting edges ordered by the IG mask versus random deletion, both for the VGAE and GIN models. We can see that while the VGAE has almost no gain, the GIN model is consistently better than the random baseline. We obtained similar results for the other explainers (not shown).

\begin{table}[h!]
    \centering
    \resizebox{1.0\textwidth}{!}{
    \begin{tabular}{|c|c||c|c||c|c||c|c|}
        \hline
        \multirow{2}{4em}{} & &
        \multicolumn{2}{c||}{Cora} &
        \multicolumn{2}{c||}{PubMed} &
        \multicolumn{2}{c|}{DDI}\\
         & & edge & feature & edge & feature & edge & feature\\
          \hline
         &GE & $0.13\pm 0.22$ & $0.86\pm 32$ & $0.28\pm 0.30$ & $0.68\pm 0.32$ & $0.93\pm 0.38$ & $0.71\pm 0.31$\\
         Insertion & IG & $\mathbf{0.72\pm 0.27}$ & $\mathbf{0.96\pm 22}$ & $\mathbf{0.89\pm 0.25}$ & $\mathbf{0.92\pm 0.24}$ & $\mathbf{0.96\pm 0.29}$ & $\mathbf{0.88\pm 0.24}$ \\
         scores & DC & $0.67\pm 0.31$ & $0.70\pm 0.27$ & $0.77\pm0.29$ & $0.65\pm 0.30$ & $0.92\pm0.27$ & $-0.12\pm 0.10$\\
         &LRP & $0.02\pm 0.32$ & $0.01\pm 0.40$ & $0.03\pm 0.40$ & $0.13\pm 0.41$ &  $-0.1\pm 0.35$ & $-0.04\pm 0.28$\\
         \hline
         \hline
         &GE & $0.08\pm 0.11$ & $0.11 \pm 0.20$ & $0.03\pm 0.10$ & $0.05\pm 0.20$ & $0.22\pm 0.12$ & $0.10\pm 0.14$\\
         Deletion & IG & $\mathbf{0.23\pm 0.11}$ & $\mathbf{0.31\pm 0.18}$ & $\mathbf{0.26\pm 0.15}$ & $\mathbf{0.26\pm 0.22}$ & $\mathbf{0.43\pm 0.15}$ & $\mathbf{0.19\pm 0.21}$\\
         scores & DC & $0.23\pm 0.11$	& $0.05\pm 0.18$ & $0.24\pm 0.15$ & $0.12\pm 0.11$ & $0.38\pm 0.13$ & $0.01\pm 0.15$\\
         &LRP & $0.01\pm 0.25$ & $0.01\pm 0.44$ & $-0.11\pm 0.38$ & $-0.36\pm 0.39$ & $0.23\pm 0.31$ & $0.03\pm 0.33$\\
         \hline
    \end{tabular}
    }

    \vspace*{8pt}
    \caption{Area scores (median $\pm$ std) for insertion (top) and deletion (bottom) for the selected explainers: GNNExplainer (GE), Integrated Gradients (IG), Deconvolution (DC) and Layer-wise Relevance Propagation (LRP). Area scores are calculated by taking into consideration the explanation scores produced by the explainers for LP models with GIN architecture as the encoder, trained with the three datasets: Cora, PubMed and DDI.}
    \label{tab:area_score_insertion_deletion}
\end{table}

\section{Discussion Of Findings}
In the previous section we devised different approaches for a quantitative comparison of explanation methods applied to the link prediction task.\footnote{The complete source code is available at \href{https://github.com/cborile/eval_lp_xai}{https://github.com/cborile/eval\_lp\_xai}} Synthetic data offers the advantage of having a ground truth available and complete control over its construction, but methods for a quantitative evaluation of real-world data, where no information is available \emph{a priori}, are also necessary. For the latter we introduced the \emph{area score}, a single-valued metric based on the insertion and deletion curves introduced in \cite{petsiuk2018rise} that quantifies the gain in performance with respect to the random baseline when node features and/or edges are inserted/removed according to the attribution scores.

IG performs better in all cases, and this is coherent with previous results on GCNs for node and graph classification tasks~\cite{sanchez2020evaluating,faber2021comparing}. Deconvolution is a good alternative. We note that GNNExplainer, despite the acceptable performance, needs to be trained, and its output is strongly dependent on the choice of its hyperparameters. This makes it difficult to use GNNExplainer as a plug-and-play method for the attribution of GNN models. It has the advantage of being model-agnostic, contrary to the other methods.

Applied to the DDI dataset, the utility of the area score and the insertion and deletion curves is particularly clear, since the drug-drug interaction graph is much more dense than the other examples. When looking for the reason of a link prediction output obtained through a Black-Box GML model, there are normally too many neighboring edges contributing to the model output even for 1- or 2- layers Graph Neural Networks, i.e., GNNs that consider only 1- or 2-hop neighborhoods in their computation graphs. A good area score on the edges means that most of the neighboring edges can be discarded for explaining the model output, thus increasing the interpretability for experts of what drugs can explain the interaction between a new candidate drug and existing ones.   

Finally, we showed that technical details of the GNN black-box models can result in very different attributions for the same learned task, and even make some explanation methods completely inapplicable. Some of these details, like the choice of the distance function in the decoder stage, are inherent for the link prediction task and must be taken carefully into account when explainability is important. Also, different graph neural network architectures can result in drastic changes in the explanations, as some architectures can weigh more the network structure, while others can extract more signal from the node features.

\section{Conclusions and Future Work}
We introduced quantitative metrics for evaluating GML model explanations in LP tasks using a synthetic dataset testbed with known ground truth and adapted insertion/deletion curves for empirical datasets. This provided metrics for validating attribution methods when ground truth is unavailable. We tested representative XAI methods on GML models with different architectures and datasets, and our metrics enabled comparison of LP explanations with each other and with random baselines.

The thorough comparison of explanations we performed revealed hidden pitfalls and unexpected behaviors. For example, we identified cases where two models with similar performance produce drastically different explanations, and how seemingly minor choices, like embedding similarity in decoders, significantly impact explanations. The integration of feature and edge explanation scores, often overlooked in GML XAI, is a promising area for future research. We strongly advocate for comparative validation of XAI techniques, enabling informed selection of explainers, and we believe that the development of validation metrics and benchmarks is the first step towards establishing quantitative, comparative validation protocols for XAI techniques.
This, in turn, would enable awareness in the choice of both GML models and explainers, and critical acceptance of the produced explanations.

Besides its technical challenges, explainable LP is a task that might positively impact several real-world scenarios, spanning from social networks, to biological networks and financial transaction networks. Each of these application domains displays unique characteristics and behaviors, both on the pragmatical and semantic level, and might therefore require the careful selection of an explainer in order to trust the final explanation. A pipeline that seamlessly integrates a GML model with an explainer, combining results of both model performance and explanation accuracy with the area score, might help mitigate the well-known black-box problems: difficulty of adoption from domain experts and debugging from developers, legal risk of non-compliance to regulation, and moral risk of inadvertently deploying biased models.
%
%
%
%
\bibliographystyle{abbrv}
\bibliography{biblio.bib}
\end{document}